\documentclass[runningheads]{llncs}
\usepackage{graphicx}
\usepackage{pbox}
\usepackage{tikz}
\usepackage{pgfgantt}
\usepackage{subfigure}
\usepackage{gensymb}
\usepackage{amsmath}
\usepackage{adjustbox} 
\usepackage{amsfonts}

\usepackage{array}
\usepackage{graphicx}
\usepackage{multirow}
\usepackage{amsfonts}
\usepackage{amssymb}
\usepackage{placeins}
\usepackage{url}
\usepackage{graphicx}
\usepackage{epstopdf}

\usepackage{algorithm}
\usepackage[noend]{algpseudocode}

\usetikzlibrary{matrix,shapes,arrows,positioning,chains}
\tikzstyle{process} = [rectangle, minimum width=3cm, minimum height=1cm, text centered, draw=black]
\tikzstyle{arrow} = [thick,->,>=stealth]
\tikzstyle{io} = [trapezium, trapezium left angle=70, trapezium right angle=110, text centered]

\DeclareMathOperator*{\argmax}{arg\,max}

\newcommand\blfootnote[1]{%
  \begingroup
  \renewcommand\thefootnote{}\footnote{#1}%
  \addtocounter{footnote}{-1}%
  \endgroup
}

\begin{document}
\title{Segmentation of VHR EO Images using Unsupervised Learning}
%
%

\author{Sudipan Saha\inst{1}\orcidID{0000-0002-9440-0720} \and
Lichao Mou\inst{1,2}\orcidID{0000-0001-8407-6413} \and
Muhammad Shahzad\inst{1}\orcidID{0000-0002-8278-9118} \and
Xiao Xiang Zhu\inst{1,2}\orcidID{0000-0001-5530-3613}}
\authorrunning{S. Saha et al.}
%
\institute{Data Science in Earth Observation, Technical University of Munich, Germany \\
\email{\{sudipan.saha,muhammad.shahzad\}@tum.de}
\and
Remote Sensing Technology Institute, German Aerospace Center (DLR), Germany \\
\email{\{lichao.mou,xiaoxiang.zhu\}@dlr.de}}
\maketitle              
\begin{abstract}
Semantic segmentation is a crucial step in many Earth observation tasks. Large quantity of pixel-level annotation is required to train deep networks for semantic segmentation. Earth observation techniques are applied to varieties of applications and since classes vary widely depending on the applications, therefore, domain knowledge is often required to label Earth observation images, impeding availability of labeled training data in many Earth observation applications. To tackle these challenges, in this paper we propose an unsupervised semantic segmentation method that can be trained using just a single unlabeled scene. Remote sensing scenes are generally large. The proposed method exploits this property to sample smaller patches from the larger scene and uses deep clustering and contrastive learning to refine the weights of a lightweight deep model composed of a series of the convolution layers along with an embedded channel attention. After unsupervised training on the target image/scene, the model automatically 
segregates the major classes present in the scene and produces the segmentation map. Experimental results on the Vaihingen dataset demonstrate the efficacy of the proposed method.


\keywords{Segmentation  \and Unsupervised learning \and Earth observation.}
\end{abstract}

\blfootnote{Accepted in ECML/PKDD workshop Machine Learning for Earth Observation (MACLEAN) 2021.}

\fboxsep=0mm
\fboxrule=0.1pt

\section{Introduction}
\label{secIntro}
Plethora of satellites equipped with High Resolution sensors have been launched in the last decade. Additionally, unmanned aerial
vehicles (UAVs) are now widely available, thus generating a large volume of  images for detailed Earth observation (EO). Automatic parsing of such images is useful for
various applications, including disaster management \cite{saha2020building} and urban monitoring \cite{saha2020unsupervised}.
The last decade also witnessed the development of various deep learning methods that outperform the previous methods on EO images.
Their superior performance is attributed to their capability to learn complex spatial features from large volume of labeled data. 
\par
A crucial step in understanding EO images is semantic segmentation, i.e., assigning a semantically meaningful class/category to each pixel in the image.
Semantic segmentation for natural images has progressed fast exploiting availability of vast training data and superior performance of convolutional neural
networks (CNNs).  While CNN based methods have been adopted to the EO images \cite{mou2020relation}, their applicability in the EO has been
limited due to the lack of labeled data \cite{hua2021semantic}.
\par
While the lack of training data offers a hurdle for segmentation of VHR EO images, their large spatial size offers an advantage. Being representative of a geographical area, EO scenes are generally large, e.g.,
each scene in the Potsdam dataset (part of ISPRS semantic labeling
dataset \cite{isprs2DSemanticLabeling}) has a size of 6000 $\times$ 6000 pixels. While a typical image in the computer vision datasets \cite{everingham2015pascal} rarely captures
multiple instances of the same object in the same image, EO images may capture 
even up to hundreds of instances of the same object (e.g., building) in a single image/scene. 
Though most state-of-the-art semantic segmentation
methods are supervised \cite{long2015fully,ronneberger2015u}, there are few methods based on the concept of
deep clustering \cite{saha2019semantic} that can work in unsupervised manner. The unsupervised paradigm has also been
extended for remote sensing images in context of multitemporal analysis by exploiting temporal consistency between images in a time-series \cite{saha2020unsupervised}. Inspired by their success,
we propose an semantic segmentation method  that can learn the segmentation clusters in unsupervised manner from a single image using a lightweight model. The proposed method employs deep clustering
and contrastive learning and produces segmentation map such that each label corresponds to a semantically meaningful entity. The proposed method is trained on a single scene only.  
The key contributions of this paper are as follows:
\begin{enumerate}
\item Proposing an unsupervised segmentation method that can be trained on a single unlabeled EO scene.
\item Incorporating deep clustering and contrastive learning in same framework for unsupervised segmentation.
\end{enumerate}
Related works are discussed in Section \ref{sectionRelatedWorks}. We define the problem statement and detail the proposed single-scene segmentation method in
Section \ref{sectionProposedMethod}. Experimental validation is presented Section \ref{sectionResults}. We conclude
the paper and discuss scope of future research in Section \ref{sectionConclusion}.

\section{Related Work}
\label{sectionRelatedWorks}
Considering relevance to our work, in this Section we briefly discuss deep segmentation, unsupervised and self-supervised learning.
\subsection{Deep segmentation}
\label{subSectionRelatedWorksDeepSegmentation}
Most supervised methods for semantic segmentation rely on pixelwise classification using a classifier that is trained using available reliable training pixels. Many deep learning based 
supervised methods have been proposed in the literature \cite{ronneberger2015u,volpi2017dense,chen2018deeplab}.
Most deep learning based approaches have an architecture consisting of an encoder and a decoder to achieve discrimination at pixel level. Several supervised segmentation methods have been proposed for EO images \cite{sherrah2016fully,volpi2017dense,maggiori2017high,marmanis2018classification,mou2020relation,ding2020semantic}. The supervised methods require a significant amount of training data.   To address the scarcity of training data,  \cite{hua2021semantic}
proposed a segmentation method that trains the model using incomplete annotations. Unsupervised deep clustering for multi-temporal EO segmentation is proposed in \cite{saha2020unsupervised}. However,
their method focuses on only two classes per target scene.

\subsection{Unsupervised and self-supervised learning}
\label{subsectionSelfSupLearningRelatedWork}
Supervised methods are limited in many applications due to the difficulty of labeling data. This has motivated  machine learning researchers
to develop unsupervised and self-supervised methods. Some works \cite{gidaris2018unsupervised,doersch2015unsupervised} use pre-text tasks like image rotation to learn unsupervised semantic feature.
Similar approaches have been adopted in EO, e.g., learning to rearranging randomly shuffled time-series images \cite{saha2020change}. In addition to pre-text tasks, some methods rely on
deep clustering by jointly learning the cluster assignment and weights of the deep network \cite{caron2018deep}. Given a  collection of unlabeled inputs, deep clustering divides them into groups
in terms of inherent latent semantics. Many variants of deep clustering exists, e.g., using convolutional autoencoder \cite{guo2017deep}.
Contrastive methods function by bringing the representation of positive pairs closer while spreading representations of negative pairs apart \cite{chen2020simple,chen2020improved,tian2020makes}. 
\cite{asano2019critical} demonstrated that the unsupervised methods learn useful semantic features even with a single-image input. 
\par
Our work is inspired from the above-mentioned works on unsupervised and self-supervised learning, especially deep clustering \cite{caron2018deep}.  Moreover similar to \cite{asano2019critical}, our
 work focuses on single scene. 

\section{Proposed method}
\label{sectionProposedMethod}
Let $X$ be a VHR EO image/scene of spatial dimension $R\times C$ pixels where $R$ and $C$ are much larger than usual image sizes in computer vision (224 $\times$ 224).
Originally we do not have label corresponding to any pixel in $X$. Our goal is to obtain segmentation map corresponding to $X$, i.e., we want to assign labels to each pixel in $X$ such that those
labels are semantically meaningful. We accomplish this by using self-supervised learning that do not require any external label.
Smaller patches of size $R'\times C'$ ($R'<R$ and $C'<C$) are extracted from $X$. Let us assume that a total  $\mathcal{B}_{total}$ patches can be extracted from $X$. One training batch
involves only a batch of $\mathcal{B}$ patches sampled from $\mathcal{B}_{total}$, denoted as   $\mathcal{X}=\{x^1,...,x^\mathcal{B}\}$.  $x^b$ is processed 
using a deep clustering loss, thus simultaneously refining the weights of the model and the segmentation map. We use a lightweight model that uses
a series of convolution layers and a channel attention. At the end of $\mathcal{I}$
epochs, the trained model can be applied to $X$ to obtain its segmentation map.  Furthermore we demonstrate that the self supervised network trained on $X$ can be directly applied to
another spatially disjoint but semantically similar scene $Z$, without requiring any further training or fine tuning. The proposed method is shown in Algorithm \ref{algorithmProposedSingleSceneSegmentation}.

\subsection{Basic architecture}
\label{sectionDeepFeatureRepresentation}
Usually the number of images used in training a deep network is in the order of tens of thousands. Compared to that the number of patches that can be used for unsupervised learning 
from a single EO image/scene is limited. Actual number is a function of $R, C, R', C'$. Considering this we design a lightweight network consisting of only few ($L$) convolution layers. Convolving the input patches through
convolution layers allow us to capture the pixelwise semantics. Successive convolution layers capture increasingly complex spatial details. The spatial size of the input images are preserved through the successive layers by exclusion of stride
or pooling operation from the architecture. Output from each convolution layer is processed through activation function (Rectified Linear Unit - ReLU) to introduce non-linearity and by Batch Normalization layer.
The weights of the network, denoted as  $\mathbb{W}^1,...,\mathbb{W}^{L}$ are initialized using a suitable initialization method and are trainable using a set of loss functions that do no require any external label. While the kernel numbers are fixed
as 64 in all layers (any other number could be chosen), last layer projects feature to $K$-dimensional space, where $K$ is an approximation of desired number of classes. In addition to the convolution layers, a channel attention mechanism is used. 
Channel attention mechanism has demonstrated potential in improving the performance of CNNs \cite{hu2018squeeze}. We apply the channel attention just before the final 1$\times$1 convolution layer.
The channel attention is designed following \cite{woo2018cbam}, i.e., by joint use of both average pooling and max-pooling. 
\par
After processing a patch $x^b$ in $\mathcal{X}$ through
the network, for each input pixel $x^b_{n}$ $(n=1,...,N)$, we obtain deep features $y^b_{n}$ of dimension $K$. The basic architecture
(showing only convolution and attention layers) is shown in Table \ref{theNetworkStructure}.

\begin{algorithm}[!t]
\caption{Self-supervised training for single-scene segmentation}
\begin{algorithmic}[1]
\State \textit{Input:} A VHR EO image/scene $X$ 
\State \textit{Output:} A lightweight model that can segment  $X$ or any other similar scene
\State Initialize $\mathbb{W}^1,...,\mathbb{W}^{L}$ 
\State Extract $\mathcal{B}_{total}$ patches from $X$
\For{$i \gets 1$ to $\mathcal{I}$}  
	\While  {all $\mathcal{B}_{total}$ patches are not sampled}
	\State Sample $\mathcal{B}$ patches from $\mathcal{B}_{total}$ patches, denoted as $\mathcal{X}=\{x^1,...,x^\mathcal{B}\}$
		\For{$j \gets 1$ to $\mathcal{J}$}   
			\For {$b \in \mathcal{B}$}
				\For {$n$-th pixel in $x^b$}
					\State Compute feature $y^b_n$
					\State Compute pseudo-label $c^b_n$
					\State Compute loss $\ell^b_{n}$
				\EndFor
				\State Compute $\mathcal{L}^b$ by considering all $n$ in $x^b$
			\EndFor
			\State Compute $\mathcal{L}$ by considering $b=1,...B$
			\State Shuffle $\mathcal{X}$ to $\mathcal{X'}$
			\State Compute contrastive loss $\mathcal{L}'$
			\State Update $\mathbb{W}^1,...,\mathbb{W}^{L}$ with $\mathcal{L}$ and $\mathcal{L}'$ 
		\EndFor
	\EndWhile
\EndFor
\end{algorithmic}
\label{algorithmProposedSingleSceneSegmentation}
\end{algorithm}

\begin{table}[h]
\center
\caption{Basic architecture of the network. Activation function and batch-normalization is excluded for sake of brevity.}
\begin{tabular}{|c|c|c|c|} 
\hline
\textbf{Layer} & \textbf{Kernel} & \textbf{Kernel size} & \textbf{Stride}\\ 
\hline
convolution & 64 & (3,3) & 1\\ 
\hline
convolution & 64 & (3,3) & 1\\ 
\hline
convolution & 64 & (3,3) & 1\\
\hline
convolution & 64 & (3,3) & 1\\
\hline
convolution & 64 & (3,3) & 1\\
\hline
Attention & NA & NA & NA\\
\hline
convolution & $K$ & (1,1) & 1 \\
\hline
\end{tabular}
\label{theNetworkStructure}
\end{table}

\subsection{Pseudo label assignment}
\label{sectionLabelAssignment}
Semantically similar inputs (in our case, pixels) generate strong activations in similar feature. Following this principle, we can assign each pixel to a label/cluster by using argmax classification. 
More specifically, label $c^b_{n}$ for an input pixel $x^b_{n}$ is estimated by selecting the feature in which $y^b_{n}$ has maximum value.
Representing the $k$-th feature of $y^b_{n}$ as $y^b_{n}(k)$,  $c^b_{n}$ is obtained as following:
\begin{equation}
c^b_{n}=\argmax_{k\in K} y^b_{n}(k)
\end{equation}
Considering that the last layer has $K$ different neurons, $c^b_{n}$ can take at most $K$ values. Thus this is equivalent to clustering with $K$ number of classes. Please note that we assign
label to each pixel for each patch in the training batch, i.e., our deep clustering process works at pixel level.

\subsection{Deep clustering}
\label{sectionDeepClustering}
Training the self-supervised network is composed of two processes, assignment of labels to each pixel, estimation of loss based on assigned labels. This process continues in iteration by
reassigning the weights and re-estimating loss. Label assignment of each pixels needs to be meaningfully refined so that semantic information of the image is captured
and  label assignment converges with iterations, performed for $\mathcal{J}$ iterations for each batch. Towards this, we compute the cross-entropy loss between
the continuous-valued deep feature representation $y^b_{n}$ and the discrete valued estimated labels  $c^b_{n}$.
\begin{equation}
\ell^b_{n}=\textrm{crossentropy}(y^b_{n},c^b_{n})
\end{equation}
In practice the loss term $\mathcal{L}$ is computed by considering all pixels in a patch and all patches in a training batch.

\subsection{Contrastive learning}
\label{sectionContrastiveLearning}
Contrastive learning is employed to encourage the network to produce dissimilar
feature for different input. Though we do not have any negative samples under the unsupervised setting, we shuffle the 
batch of patches $\mathcal{X}$ to $\mathcal{X'}$. This implies that $b-$th patch in $\mathcal{X}$ ($x^b$) and $\mathcal{X'}$ ($x^{b'}$) are unpaired. Thus we compute negative absolute error loss for each
input pixel $x^b_{n}$ and $x^{b'}_{n}$:
\begin{equation}
\ell^{b'}_{n}= -||(y^b_{n}-y^{b'}_{n})||_{_1}
\end{equation}
Loss term $\mathcal{L}'$ is computed  as
mean of exponentials of $\ell^{b'}_{n}$ over all considered pixels for all patches in the batch.
\par
The sum of loss term $\mathcal{L}$ and $\mathcal{L}'$  is used to update the model weights $\mathbb{W}^1,...,\mathbb{W}^{L}$.
Note that the computation of $\mathcal{L}$ does not require any external label and hence the mechanism is unsupervised.

\section{Experiments}
\label{sectionResults}
\subsection{Dataset}
We use the Vaihingen dataset that  is a benchmark dataset for semantic
segmentation provided by the International Society for Photogrammetry
and Remote Sensing (ISPRS) \cite{isprs2DSemanticLabeling}. The images are collected over the city of Vaihingen with a spatial resolution of 9 cm/pixel. Each image in the dataset convers an average area of
1.38 square km. Three bands are available - near infrared (NIR), red (R), and green (G). Additionally digital surface models (DSMs) are available that are not used in this work.
In total, six land-cover classes are considered: impervious surface, building, low vegetation,
tree, car, and clutter/background. As used in \cite{hua2021semantic}, we use image IDs 11, 15, 28, 30, and 34 as test set. Since we need only a single scene for training, image ID 1 is used for training the unsupervised model.
Our result is shown as an average of three runs with different seeds.

\subsection{Comparison method}
To the best of our knowledge, our work is first attempt to obtain multi-class segmentation maps from VHR images in unsupervised manner. Hence, comparison to supervised paradigms is unfair and instead comparison 
needs to be performed with methods that can work in label-constrained manner. Considering this, we compared the proposed method to FEature and Spatial relaTional regulArization
(FESTA) \cite{hua2021semantic} that trains semantic segmentation model based on incomplete annotations.
For comparison, we trained the FESTA model in \cite{hua2021semantic} using image ID 1 and using different number of training points. Please note that inspite of working on the incomplete annotations, the method in \cite{hua2021semantic} has access
to some labeled point during training, while the proposed method does not use any annotated data during training.
 
\subsection{Result}
The training process is accomplished with $\mathcal{I} = 2$,  $\mathcal{J} = 50$, and number of kernels in the final layer $K =8$, a value considerably close to number of classes in the images (6).
Choice of $K =8$ is the only component of the proposed method, where prior knowledge about the target scene is used. Since our approach is unsupervised, it is not possible to
automatically know the name of each class unlike supervised segmentation. Here we have assigned each class a name as per their overlap with the classes in the reference map. 
\par
\begin{figure*}[!h]
\centering
{\fcolorbox{black}{white}{\rule{0pt}{6pt}\rule{12pt}{0pt}}\quad Impervious surfaces}\hspace{10 pt} 
{\fcolorbox{blue}{blue}{\rule{0pt}{6pt}\rule{12pt}{0pt}}\quad Buildings}\hspace{10 pt}
{\fcolorbox{cyan}{cyan}{\rule{0pt}{6pt}\rule{12pt}{0pt}}\quad Low vegetation}\hspace{10 pt}
\\
{\fcolorbox{green}{green}{\rule{0pt}{6pt}\rule{12pt}{0pt}}\quad Trees}\hspace{10 pt} 
{\fcolorbox{yellow}{yellow}{\rule{0pt}{6pt}\rule{12pt}{0pt}}\quad Cars}\hspace{10 pt}
{\fcolorbox{red}{red}{\rule{0pt}{6pt}\rule{12pt}{0pt}}\quad Clutter}\hspace{10 pt}
{\fcolorbox{black}{black}{\rule{0pt}{6pt}\rule{12pt}{0pt}}\quad Undefined}\hspace{10 pt}
\\
\subfigure[]{%
            
         \fbox{\includegraphics[height=4.8 cm]{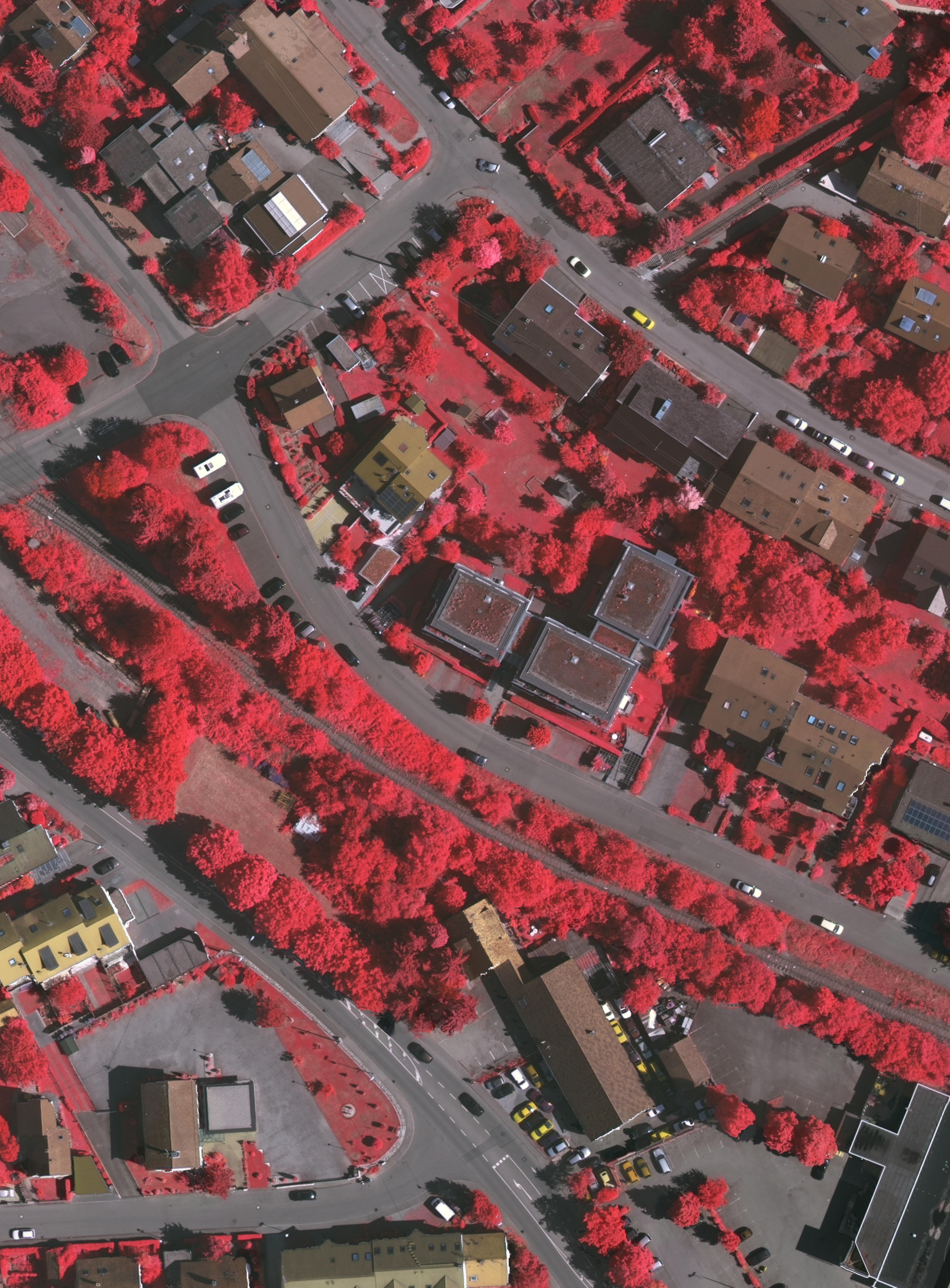}}
            \label{figureVaihingen_11_Input}
        }%
\hspace{0.2 cm}
\subfigure[]{%
            
         \fbox{\includegraphics[height=4.8 cm]{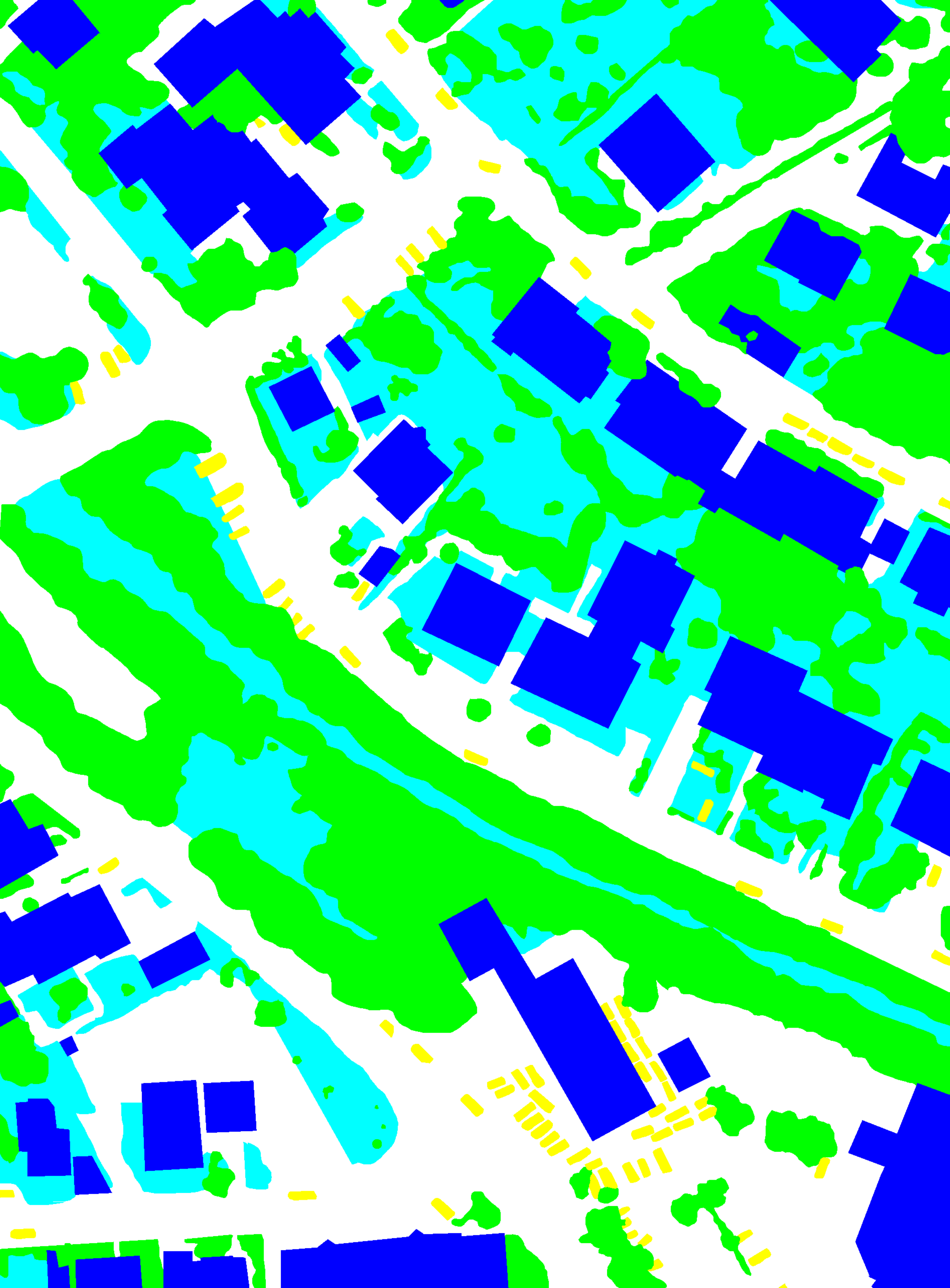}}
            \label{figureVaihingen_11_Reference}
        }%
\hspace{0.2 cm}
  \subfigure[]{%
 \fbox{\includegraphics[height=4.8 cm]{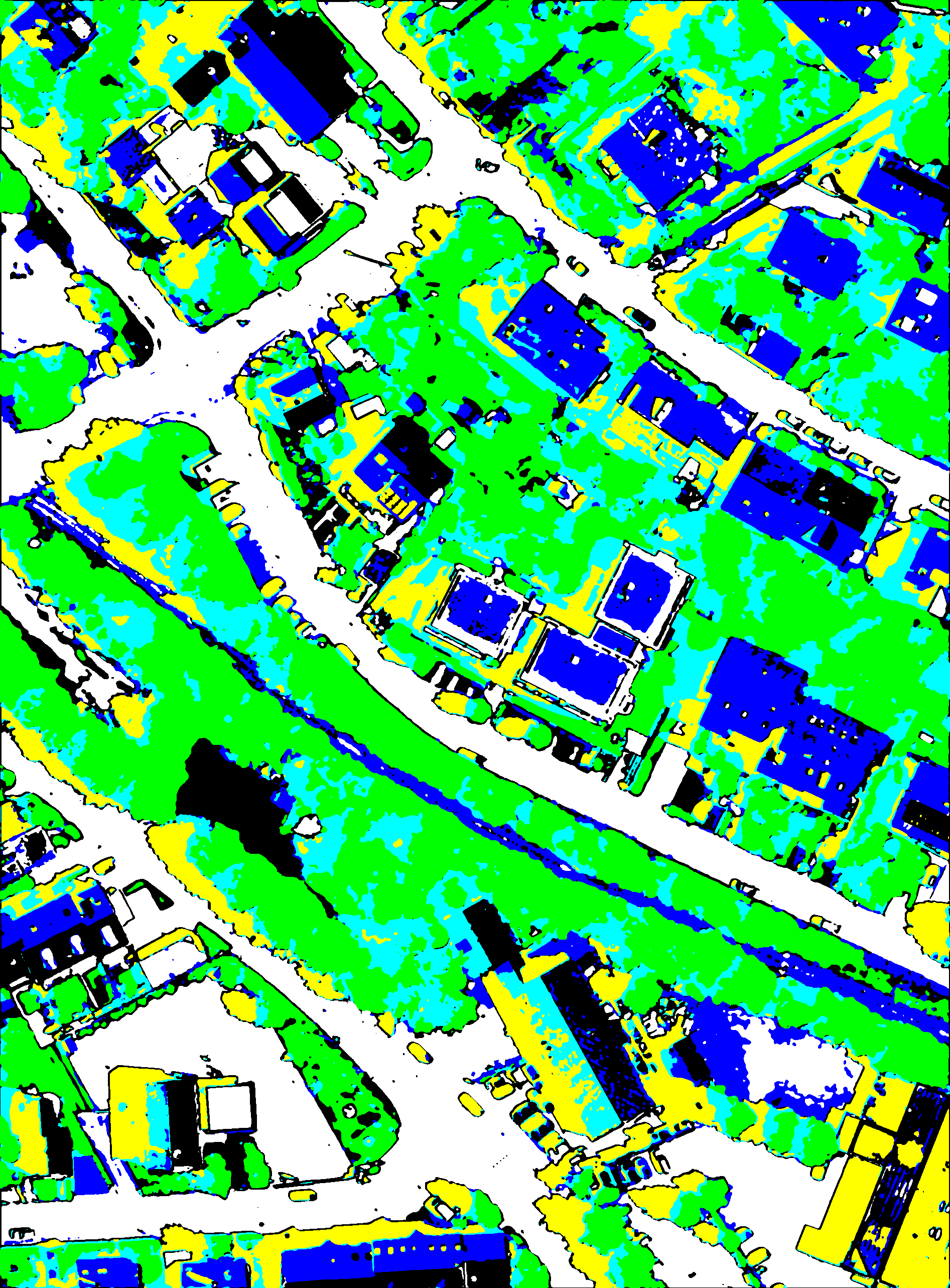}}
            \label{figureVaihingen_11_SegmentationProposed}
        }%
\\
\subfigure[]{%
            
         \fbox{\includegraphics[height=4.8 cm]{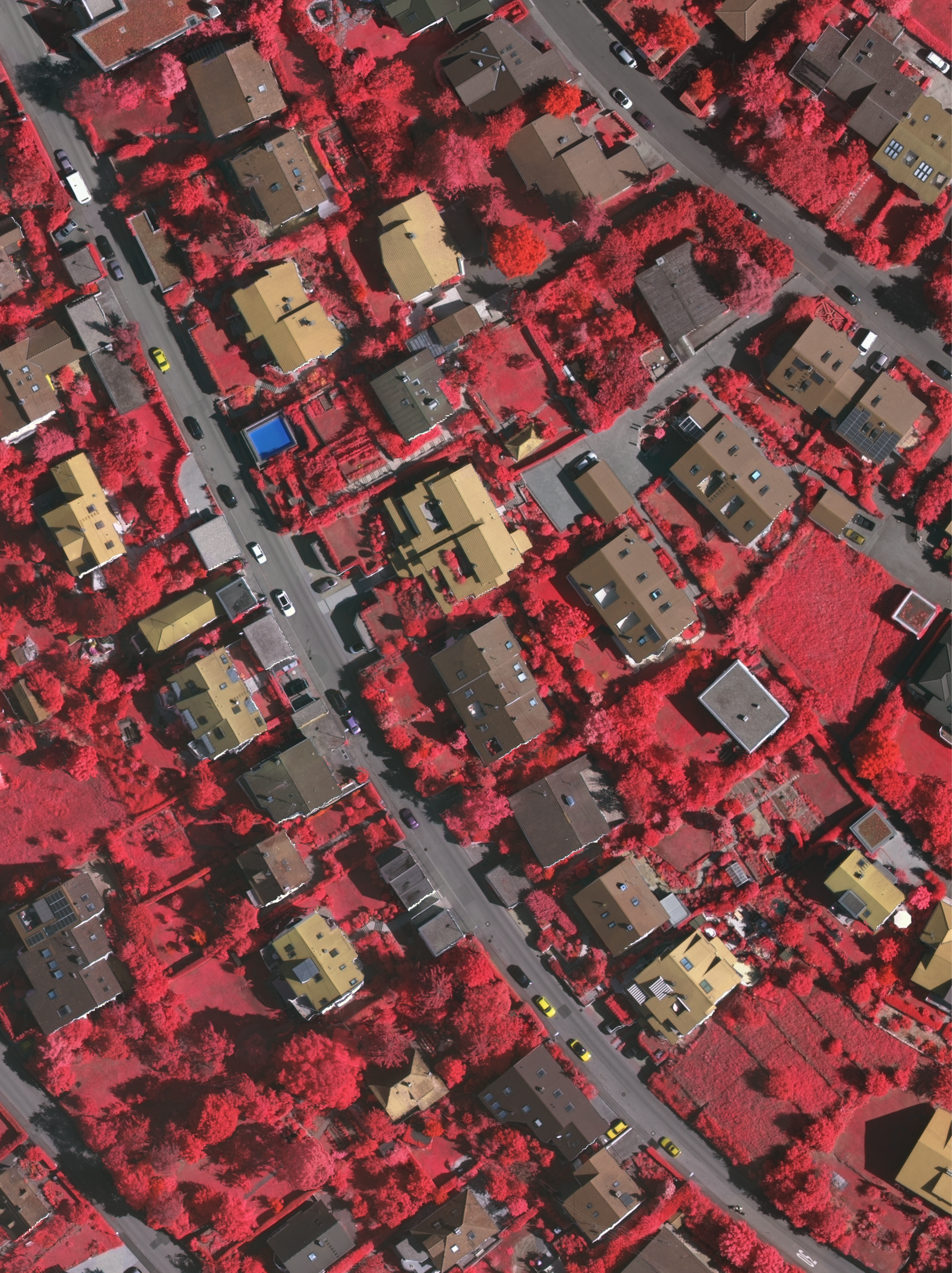}}
            \label{figureVaihingen_15_Input}
        }%
\hspace{0.2 cm}
\subfigure[]{%
            
         \fbox{\includegraphics[height=4.8 cm]{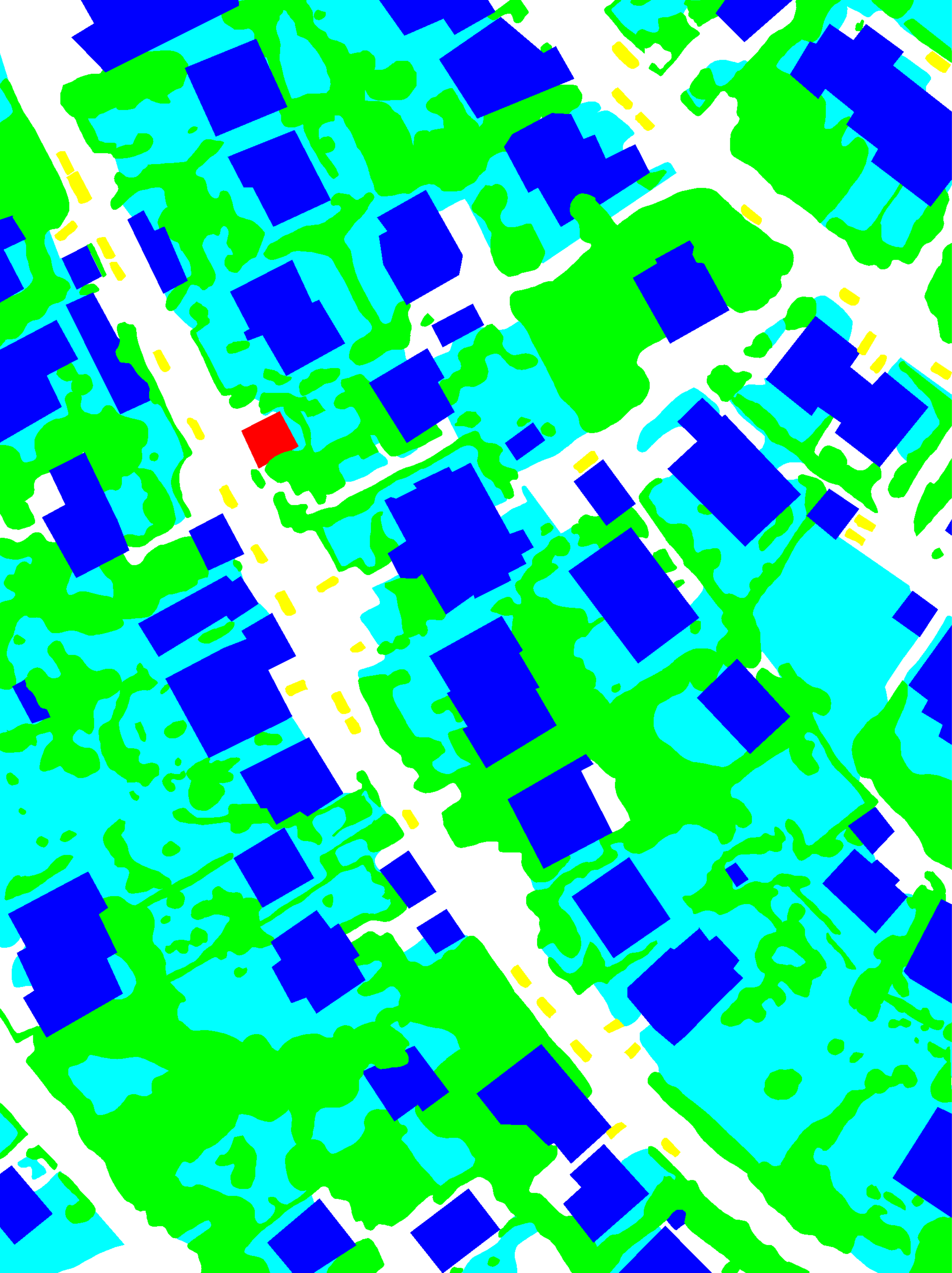}}
            \label{figureVaihingen_15_Reference}
        }%
\hspace{0.2 cm}
  \subfigure[]{%
 \fbox{\includegraphics[height=4.8 cm]{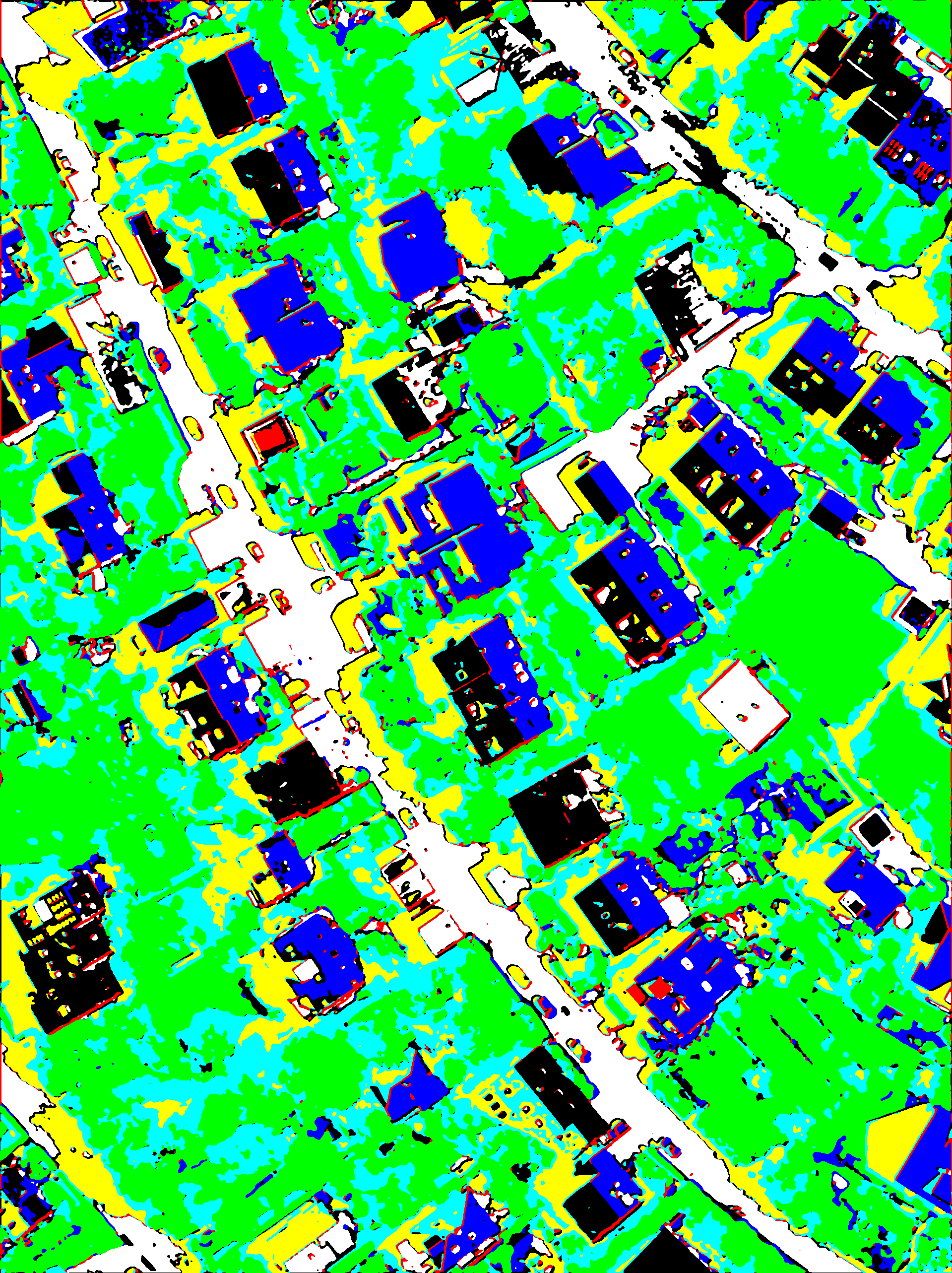}}
            \label{figureVaihingen_15_SegmentationProposed}
        }%

\caption{Visualization of segmentation on Vaihingen dataset: (a), (d) input images 11 and 15 (false color composition), (b), (e) corresponding reference segmentation, and (c), (f) are segmentation produced by the proposed unsupervised method.}
\label{figureSegmentationVaihingen}
\end{figure*}

\textbf{Multi-class segmentation:} We show the result for image ID 11 and 15 in Figure \ref{figureSegmentationVaihingen}. Input images are shown in first column. Second and third columns show the reference segmentation masks and the result obtained by proposed method, respectively. We observe that in both cases the two major classes - buildings (blue) and impervious surfaces (white) are satisfactorily detected. A significant overlap is observed between low vegetation (cyan) and trees (green), especially in image ID 15. Considering the unsupervised nature of the proposed method, it is difficult for it to know the real class divisions as desired in the reference map. Thus it identifies similar (as per spectral characteristics) low vegetation and trees as same class. 
\par
The quantitative result averaged over 5 test tiles are shown in Table \ref{tableResultVaihingen} in terms of F1 score and Intersection-Over-Union (IoU). The proposed method clearly outperforms FESTA  \cite{hua2021semantic} for 5 point and 20 point annotations.
This result shows that proposed method, despite not using any annotated data during training, can outperform existing state-of-the-art method when using few annotated points. When FESTA uses all
annotated points in tile 1, the proposed method still outperforms FESTA, however the margin reduces. 
\par
\begin{figure*}[!h]
\centering
\subfigure[]{%
            
         \fbox{\includegraphics[height=4.5 cm]{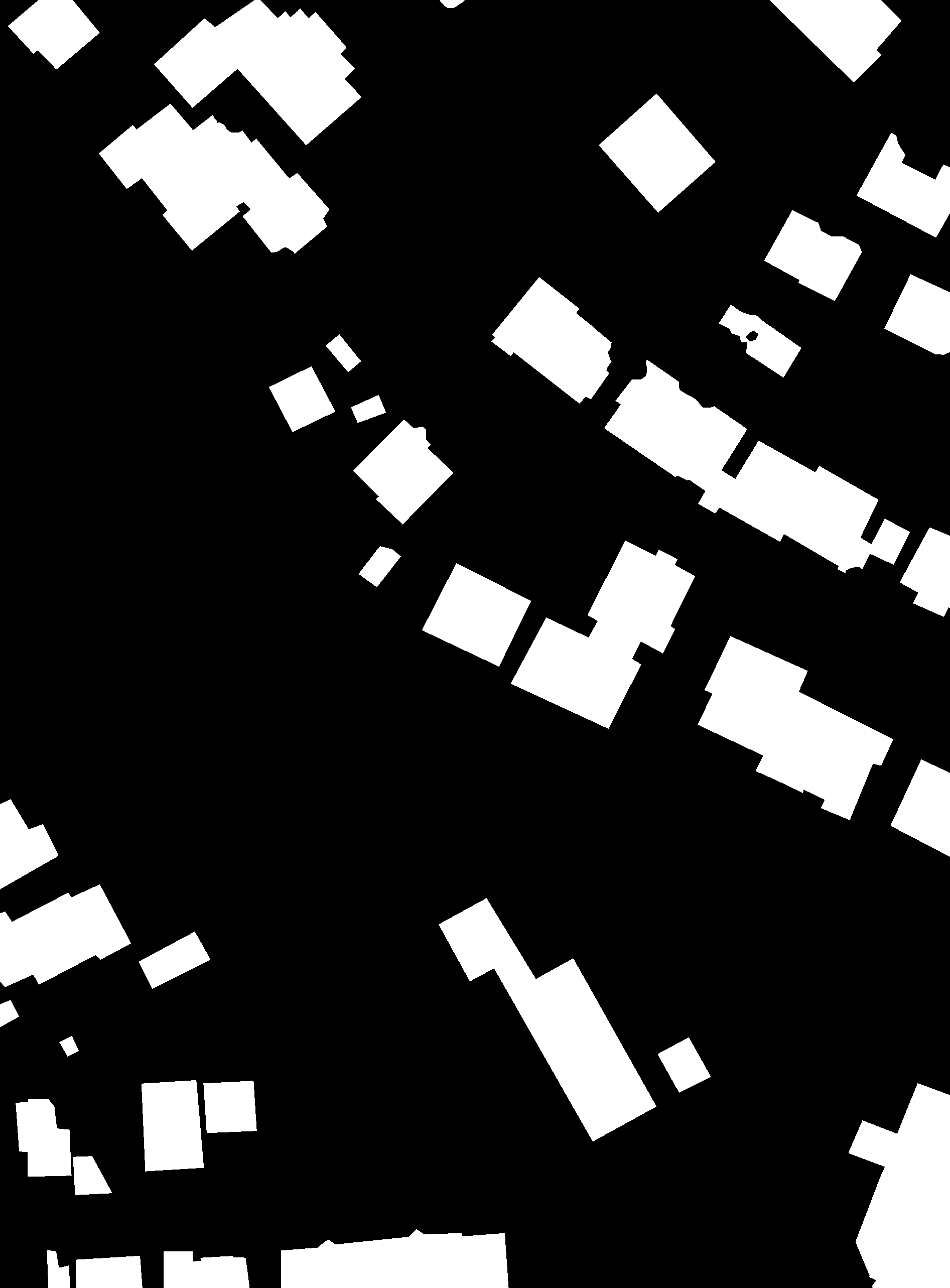}}
            \label{figureVaihingen__11_ReferenceBinary}
        }%
\hspace{0.2 cm}
\subfigure[]{%
            
         \fbox{\includegraphics[height=4.5 cm]{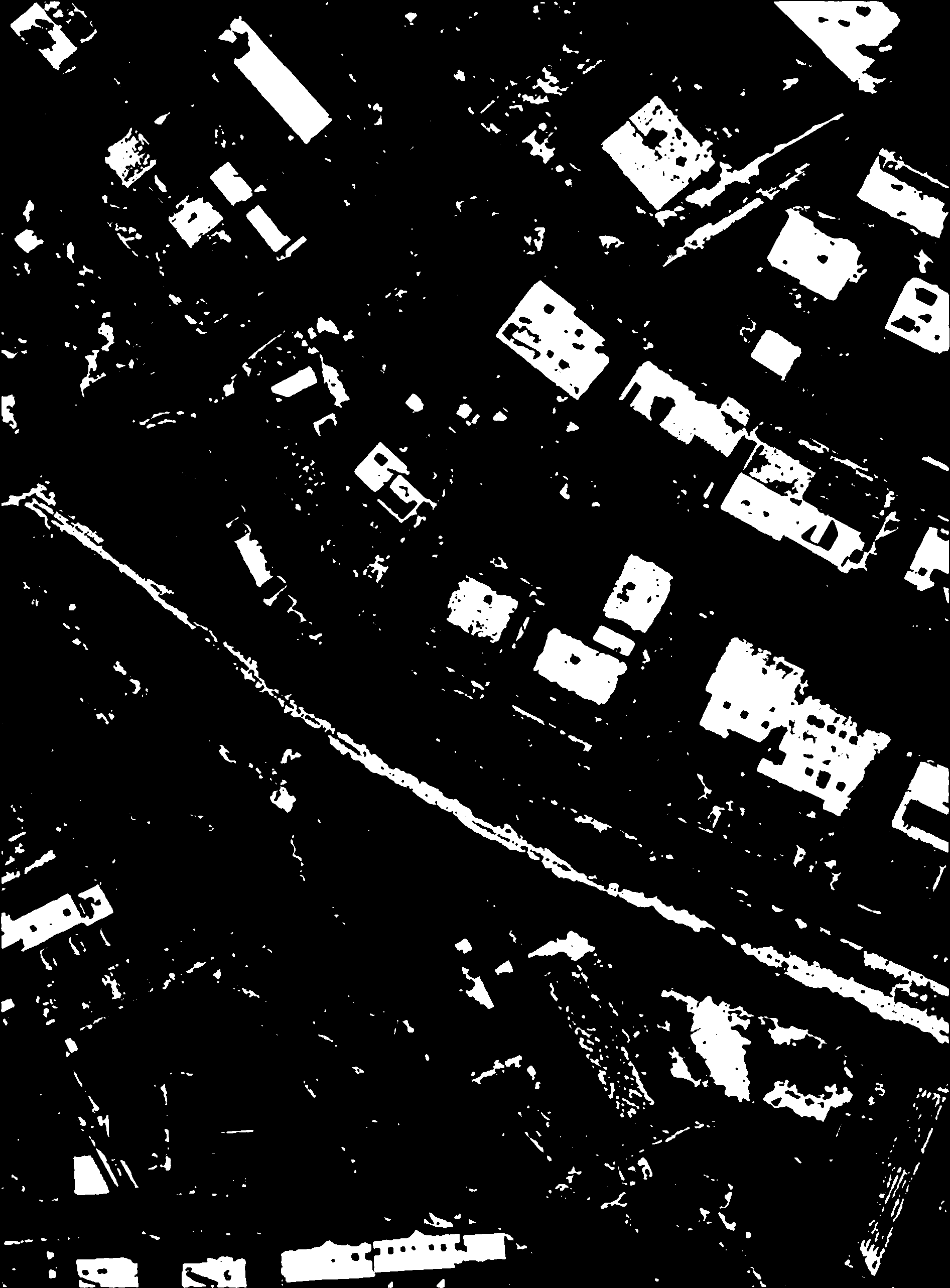}}
            \label{figureVaihingen__11_SegBinary}
        }%

\caption{Visualization of binary (building/other classes) segmentation on Vaihingen dataset image ID 11: (a) reference  11 , (b) segmentation produced by the proposed method.}
\label{figureSegmentationVaihingenBinary}
\end{figure*}

\textbf{Binary segmentation:} In many urban applications, it is more important to know only the man-made urban structures than low vegetation and trees. Considering that, we also show the performance of the proposed method as
binary segmentation map, considering two classes: buildings  as one class (white) and rest as one (black). For image 11, Figure \ref{figureVaihingen__11_ReferenceBinary} shows the 
reference binary segmentation map and Figure \ref{figureVaihingen__11_SegBinary} shows the binary segmentation obtained by the proposed method. It is visually evident that there is high match between the binary reference map and
the binary segmentation map.

\renewcommand{\tabcolsep}{2pt}
\begin{table}
\centering
\caption{Quantitative comparison of the proposed method to FESTA \cite{hua2021semantic}.}
\begin{tabular}{|c|c|c|} 
 \hline
\textbf{Method} & \textbf{F1 score} & \textbf{IOU}   \\ 
\hline
\bf Proposed Unsupervised  & 0.43 & 0.30 \\ 
\hline
FESTA 5 points   & 0.26 & 0.16  \\ 
\hline
FESTA 10 points  & 0.32 & 0.23\\ 
\hline
FESTA All points  & 0.41 & 0.28 \\ 
\hline
  \end{tabular}
\label{tableResultVaihingen}
\end{table}

\section{Conclusions}
\label{sectionConclusion}
  This paper proposed a deep clustering and contrastive learning based unsupervised semantic segmentation method for single scene EO images. Exploiting the large spatial 
  size of the EO images, the proposed method divides the image into patches that are
  further used for training the unsupervised network. Pseudo labels 
  are obtained by argmax classification of the final layer. The proposed method optimizes the labels and weights in iterations. The experimental results on Vaihingen dataset show the efficacy of the proposed method to obtain meaningful
  segmentation labels. 
 Instead of seen as a competitor, the proposed method should be seen as a complementary to the existing supervised segmentation methods in EO. Since the proposed method provides a fast way to predict reasonably accurate
 segmentation map using single scene, it may be useful in conjunction with supervised methods  to generate pseudo-labels. Though we applied the proposed method to urban scenes, the model is application-agnostic. Our future work
  will aim towards improving the unsupervised segmentation of smaller classes (e.g., cars) and extending the proposed approach for ingesting other sensors, e.g., VHR Synthetic Aperture Radar (SAR) sensor.

%
%
%
\bibliographystyle{splncs04}
\bibliography{unsupSegmentationMaclean}

\end{document}